# SleePyCo: Automatic Sleep Scoring with Feature Pyramid and Contrastive Learning

Seongju Lee, Yeonguk Yu, Seunghyeok Back, Hogeon Seo, and Kyoobin Lee, *Member, IEEE*

**Abstract**—Automatic sleep scoring is essential for the diagnosis and treatment of sleep disorders and enables longitudinal sleep tracking in home environments. Conventionally, learning-based automatic sleep scoring on single-channel electroencephalogram (EEG) is actively studied because obtaining multi-channel signals during sleep is difficult. However, learning representation from raw EEG signals is challenging owing to the following issues: 1) sleep-related EEG patterns occur on different temporal and frequency scales and 2) sleep stages share similar EEG patterns. To address these issues, we propose a deep learning framework named *SleePyCo* that incorporates 1) a feature pyramid and 2) supervised contrastive learning for automatic sleep scoring. For the feature pyramid, we propose a backbone network named *SleePyCo-backbone* to consider multiple feature sequences on different temporal and frequency scales. Supervised contrastive learning allows the network to extract class discriminative features by minimizing the distance between intra-class features and simultaneously maximizing that between inter-class features. Comparative analyses on four public datasets demonstrate that *SleePyCo* consistently outperforms existing frameworks based on single-channel EEG. Extensive ablation experiments show that *SleePyCo* exhibits enhanced overall performance, with significant improvements in discrimination between the N1 and rapid eye movement (REM) stages.

## 1 INTRODUCTION

SLEEP scoring, also referred to as "sleep stage classification" or "sleep stage identification," is critical for the accurate diagnosis and treatment of sleep disorders [1]. Individuals suffering from sleep disorders are at risk of fatal complications such as hypertension, heart failure, and arrhythmia [2]. In this context, polysomnography (PSG) is considered the gold standard for sleep scoring and is used in the prognosis of typical sleep disorders (e.g., sleep apnea, narcolepsy, and sleepwalking) [3]. PSG consists of the biosignals associated with bodily activities such as brain activity (electroencephalogram, EEG), eye movement (electrooculogram, EOG), heart rhythm (electrocardiogram, ECG), and chin, facial, or limb muscle activity (electromyogram, EMG). Generally, experienced sleep experts examine these recordings based on sleep scoring rules and assign 20- or 30-s segments of the PSG data (called "epoch") to a sleep stage. Rechtschaffen and Kales (R&K) [4] and American Academy of Sleep Medicine (AASM) [5] standards serve as typical sleep scoring rules. The R&K standard classifies sleep stages into wakefulness (W), rapid eye movement (REM), and non-REM (NREM). NREM is further subdivided into S1, S2, S3, and S4 or N1, N2, N3, and N4. In the AASM rule, N3 and N4 are merged into N3, and it categorizes the PSG epochs into five sleep stages. Recently, the improved version of R&K—the AASM rule—has been widely utilized in manual sleep scoring. According to this rule, sleep experts should visually analyze and categorize the epochs of the entire night to form a hypnogram. Thus, manual sleep scoring is an arduous and time-consuming process [6]. By contrast, machine learning algorithms require less than a few minutes for sleep scoring [7], and their performance is comparable to that of sleep experts [8]. Therefore, automatic sleep scoring is highly desired for fast and accurate healthcare systems.

Several methods have been developed for automatic sleep scoring based on deep neural networks. Basic one-to-one schemes that utilize a single EEG epoch and produce its corresponding sleep stage, were proposed as early methods [9], [10]. In addition, one-to-many (*i.e.*, multitask learning) [11] schemes have been presented. Because the utilization of multiple EEG epochs offers advantageous performance, several many-to-one methods [12], [13], [14], [15], the method of predicting the sleep stage of the target epoch with the given PSG signals, and many-to-many (*i.e.*, sequence-to-sequence) [7], [16], [17] methods have been proposed for automatic sleep scoring. The existing methods are generally based on convolutional neural networks (CNNs) [9], [11], [12], [13], [14], [18], [19], recurrent neural networks (RNNs) [17], deep belief networks (DBNs) [20], convolutional recurrent neural networks (CRNNs) [7], [15], [16], [21], [22], [22], [23], [24], fully convolutional networks (FCNs) [25], [26], transformer [27], CNN+Transformer [10], [28], and other network architectures [29], [30].

To obtain an improved representation from EEG, the architectures are designed to extract multiscale features with varying temporal and frequency scales [9], [16], [22], [28], [30], [31], [32]. Phan *et al.* [9] and Qu *et al.* [28] used two distinct widths of max-pooling kernels on the spectrogram. Supratak *et al.* [16], Fiorillo *et al.* [31], and Huang *et al.* [30] designed two-stream networks with two distinct filter

- Seongju Lee, Yeonguk Yu, Seunghyeok Back, and Kyoobin Lee are with the School of Integrated Technology, Gwangju Institute of Science and Technology, Gwangju, Republic of Korea, 61005.
- Hogeon Seo is with Korea Atomic Energy Research Institute, Daejeon, Republic of Korea, 34057.
- Kyoobin Lee (kyoobinlee@gist.ac.kr) is the corresponding author.

Source code is available at https://github.com/gist-ailab/SleePyCo.

widths of the convolutional layer in representation learning. Further, Sun *et al.* [22] and Wang *et al.* [32] utilized convolutional layers with two or more distinguished filter widths in parallel. These studies utilized feature maps with different receptive field sizes to obtain richer information from given input signals. However, they could not obtain the advantages of multi-level features, which represent broad temporal scales and frequency characteristics.

Automatic scoring methods based on batch contrastive approaches have been proposed [33], [34], [35] to improve the representation of PSG signals without labeled data, as multiple self-supervised contrastive learning frameworks have been proposed for visual representation [36], [37], [38]. These batch-based approaches have been extensively studied because they outperform the traditional contrastive learning methods [39] such as the triplet [40] and N-pair [41] strategies. Mohsenvand *et al.* [33] proposed self-supervised contrastive learning for electroencephalogram classification. Jiang *et al.* [34] proposed self-supervised contrastive learning for EEG-based automatic sleep scoring. CoSleep [35] presented self-supervised learning for multiview EEG representation between the raw signal and spectrogram for automatic sleep scoring. These studies only solve the problem of lack of labeled PSG data and do not focus on accurate automatic scoring. Furthermore, they do not leverage the large amount of annotated PSG data.

To address the aforementioned limitations, we propose *SleePyCo*, which jointly utilizes a feature pyramid [42] and supervised contrastive learning [43] to realize automatic sleep scoring. In *SleePyCo*, the feature pyramid considers broad temporal scales and frequency characteristics by utilizing multi-level features. The supervised contrastive learning framework allows the network to extract class discriminative features by minimizing the distance of intra-class features and maximizing that of inter-class features at the same time. To verify the effectiveness of the feature pyramid and supervised contrastive learning, we conducted an ablation study on the Sleep-EDF [44], [45] dataset that showed that *SleePyCo* exhibits improved overall performance in automatic sleep scoring by enhancing the discrimination between the N1 and REM stages. The results of extensive experiments and comparative analyses conducted on four public datasets further corroborate the performance of *SleePyCo*. *SleePyCo* achieves state-of-the-art performance by exploiting the representation power of the feature pyramid and supervised contrastive learning. The main contributions of this study are summarized as follows:

- We present a novel framework named *SleePyCo* that jointly utilizes a feature pyramid and supervised contrastive learning for automatic sleep scoring.
    - We incorporate the feature pyramid for automatic sleep scoring and propose the *SleePyCo-backbone* to consider various temporal and frequency scales of raw single-channel EEG.
    - We propose supervised contrastive learning to reduce the ambiguity of sleep stages by extracting class discriminative features.
- We demonstrate that *SleePyCo* outperforms the state-of-the-art frameworks on four public datasets via comparative analyses.
- We verified through an ablation study that the feature pyramid and supervised contrastive learning show synergies. Further, we analyzed the results from the perspective of sleep scoring criteria.

## 2 MODEL ARCHITECTURE

### 2.1 Problem Statement

The proposed model is designed to classify $L$ successive single-channel EEG epochs into sleep stages for the $L$-th input EEG epoch (called the target EEG epoch). Formally, $L$ successive single-channel EEG epochs are defined as $\mathbf{X}^{(L)} \in \mathbb{R}^{3000L \times 1}$, and the corresponding ground truth is denoted as $\boldsymbol{y}^{(L)} \in \{0,1\}^{N_c}$, where $\boldsymbol{y}^{(L)}$ denotes the one-hot encoding label of the target EEG epoch and $\sum_{j=1}^{N_c} y_j^{(L)} = 1$. $N_c$ indicates the number of classes and was set as 5, following the five-stage sleep classification in the AASM rule [5]. $\mathbf{X}^{(L)}$ can be described as $\mathbf{X}^{(L)} = \{\boldsymbol{x}_1, \boldsymbol{x}_2, ..., \boldsymbol{x}_L\}$, where $\boldsymbol{x}_i \in \mathbb{R}^{3000 \times 1}$ is a 30-second single EEG epoch sampled at 100 Hz.

### 2.2 Model Components

The network architecture of *SleePyCo* was inspired by *IITNet* [15]. As reported in [15], considering the temporal relations between EEG patterns in intra- and inter-epoch levels is important for automatic sleep scoring because sleep experts analyze the PSG data in the same manner. However, EEG patterns exhibit various frequencies and temporal characteristics. For instance, the sleep spindles in N2 occur in the frequency range of 12–14 Hz between 0.5–2 s, whereas the slow wave activity in N3 occur in the frequency range of 0.5–2 Hz throughout the N3 stage. To address this, we incorporated a feature pyramid into our model to enable multiscale representation because a feature pyramid can consider various temporal scales and frequency characteristics. This is based on the rationale that a feature pyramid considers various frequency components and spatial scales, such as shape and texture, in computer vision [46].

The proposed *SleePyCo* model comprises three major components: backbone network, lateral connections, and classifier network. The backbone network extracts feature sequences from raw EEG signals with multiple temporal scales and feature dimensions (number of channels). Thus, we designed a shallow network based on previous studies because they achieved state-of-the-art performance [7], [23], [25]. The lateral connections transform feature sequences with different feature dimensions into the same feature dimension via a single convolutional layer, resulting in a feature pyramid. For the classifier, a transformer encoder is employed to capture the sequential relations of EEG patterns on different temporal and frequency scales at sub-epoch levels.

#### 2.2.1 Backbone Network

To facilitate the feature pyramid, we propose a backbone network, named *SleePyCo-backbone*, containing five convolutional blocks, as proposed in [15], [25]. Each convolutional block is formed by the repetition of unit convolutional layers in the sequence of 1-D convolutional layer, 1-D batch normalization layer, and parametric rectified linear unit (PReLU) [47]. All convolutional layers have a filter width of 3, stride length of 1, and padding size of 1 to maintain the feature length within the same convolutional block. Max-pooling is performed between convolutional blocks to reduce the length of feature sequences. Additionally, a squeeze and excitation module [48] is included before the last activation function (PReLU) of each convolutional block. The details of the parameters, such as filter size, number of channels, and max-pooling size, are presented in Section 4.3.

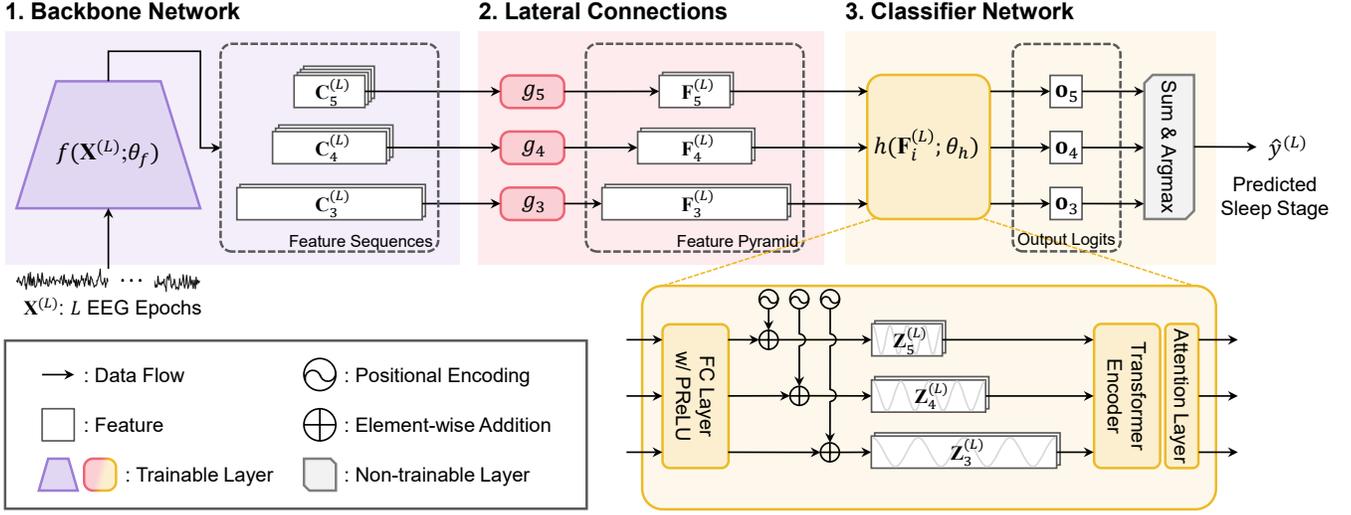

Figure 1: Model Architecture of *SleePyCo*. The purple, pink, and yellow regions indicate the backbone network, lateral connections, classifier network of *SleePyCo*, and their corresponding outputs, respectively.

Formally, *SleePyCo-backbone* takes $L$ successive EEG epochs as input, obtaining the following set of feature sequences:

$$\{\mathbf{C}_3^{(L)}, \mathbf{C}_4^{(L)}, \mathbf{C}_5^{(L)}\} = f(\mathbf{X}^{(L)}; \theta_f), \qquad (1)$$

where $\mathbf{C}_i^{(L)}$ denotes the output of the $i$-th convolutional block of the backbone network, $f(\cdot)$ represents the backbone network, and $\theta_f$ indicates its trainable parameters. The size of the feature sequence can be denoted as $\mathbf{C}_i^{(L)} \in \mathbb{R}^{\lceil 3000L/r_i \rceil \times c_i}$, where $i \in \{3, 4, 5\}$ represents the stage index of the convolutional block, $c_i \in \{192, 256, 256\}$ denotes the feature dimension, $r_i \in \{5^2, 5^3, 5^4\}$ denotes the temporal reduction ratio, and $\lceil \cdot \rceil$ signifies the ceiling operation. Notably, the temporal reduction ratio $r_i$ is derived from the ratio of the length of the input $\mathbf{X}^{(L)}$ to that of the feature sequence $\mathbf{C}_i^{(L)}$. The feature sequences from the first and second convolutional blocks were excluded from the feature pyramid owing to their large memory allocation. Thus, the stage indices of 1 and 2 were not considered in this study.

#### 2.2.2 Lateral Connections

Lateral connections were attached at the end of the 3rd, 4th, and 5th convolutional blocks to form pyramidal feature sequences (*i.e.*, feature pyramid) with identical feature dimensions. Importantly, the feature dimensions of the feature pyramid should be identical because all pyramidal feature sequences share a single classifier network. Because the feature vectors in the feature pyramid represent an assorted frequency meaning but the same semantic meaning (EEG patterns), the application of a shared classifier is appropriate in our methods. Formally, the feature pyramid $\{\mathbf{F}_3^{(L)}, \mathbf{F}_4^{(L)}, \mathbf{F}_5^{(L)}\}$ can be obtained using the following equation:

$$\mathbf{F}_i^{(L)} = g_i(\mathbf{C}_i^{(L)}; \theta_{g,i}), \qquad (2)$$

where $g_i(\cdot)$ denotes the lateral connection for the feature sequence $\mathbf{C}_i^{(L)}$ with the trainable parameter $\theta_{g,i}$. Each lateral connection consists of one convolutional layer with a filter width of 1 and results in a pyramidal feature sequence $\mathbf{F}_i^{(L)}$ that describes the same temporal scale with $\mathbf{C}_i^{(L)}$. Thus, the size of the pyramidal feature sequences is formulated as $\mathbf{F}_i^{(L)} \in \mathbb{R}^{\lceil 3000L/r_i \rceil \times d_f}$. In this study, the feature dimension $d_f$ was fixed at 128.

#### 2.2.3 Classifier Network

The classifier network of *SleePyCo* can analyze the temporal context in the feature pyramid and output the predicted sleep stage of the target epoch $\hat{y}^{(L)}$. Formally, we denote the classifier network as $h(\mathbf{F}_i^{(L)}; \theta_h)$, where $\theta_h$ is the trainable parameter. We utilized the encoder part of the Transformer [49] for sequential modelling of the feature pyramid extracted from the raw single-channel EEG. Overall, recurrent architectures such as LSTM [50] and GRU [51] were widely adopted in automatic sleep scoring. Interestingly, the transformer delivered a remarkable performance in various sequential modeling tasks, including automatic sleep scoring [27], [28], [52]. Owing to the large number of parameters of the original transformer, we reduced the model dimension $d_m$ (*i.e.*, dimension of the queue, value, and query) and the feed-forward network dimension $d_{FF}$ in comparison to the original ones. We used the original configuration of the number of heads $N_h$ and the number of encoder layers $N_e$. The parameters are detailed in Section 4.3.

Prior to the transformer encoder, a shared fully connected layer with PReLU is employed to transform the feature dimension of the feature pyramid into the dimen-sion of the transformer encoder. Specifically, we denote the transformed feature pyramid as $\tilde{\mathbf{F}}_i^{(L)} \in \mathbb{R}^{\lceil 3000L/r_i \rceil \times d_m}$. This layer maps EEG patterns from various convolutional stages into the same feature space, and thus, the shared classifier considers the temporal context, regardless of the feature level. Subsequently, because the transformer encoder is a recurrent-free architecture, the positional encoding $\mathbf{P}_i^{(L)} \in \mathbb{R}^{\lceil 3000L/r_i \rceil \times d_m}$ should be added to the input feature sequences to blend the temporal information:

$$\mathbf{Z}_i^{(L)} = \tilde{\mathbf{F}}_i^{(L)} + \mathbf{P}_i^{(L)}, \tag{3}$$

where $\mathbf{P}_i^{(L)} \in \mathbb{R}^{\lceil 3000L/r_i \rceil \times d_m}$ denotes the positional encoding matrix for the $i$-th feature sequence. Herein, sinusoidal positional encoding was performed following a prior study [49]. However, because the same time indices are applicable at both ends of the pyramidal feature sequences, we modified the positional encoding to coincide with the absolute temporal position between them by hopping the temporal index of positional encoding. Thus, the element of $\mathbf{P}_i^{(L)}$ at the temporal index $t$ and dimension index $k$ can be defined as

$$\mathbf{P}_i^{(L)}(t,k) = \begin{cases} \sin\left(\dfrac{tR^{(i-3)} + \lfloor R^{(i-3)}/2 \rfloor}{10000^{k/d_m}}\right), & \text{if } k \text{ is even,} \\ \cos\left(\dfrac{tR^{(i-3)} + \lfloor R^{(i-3)}/2 \rfloor}{10000^{k/d_m}}\right), & \text{otherwise,} \end{cases} \tag{4}$$

where $\lfloor \cdot \rfloor$ denotes the floor operation and $R = r_i/r_{i-1}$ indicates a temporal reduction factor (set as 5 in this study). $\mathbf{P}_i^{(L)}(t,k)$ degenerates into the original sinusoidal positional encoding for $i = 3$.

The output hidden states $\mathbf{H}_i^{(L)} \in \mathbb{R}^{\lceil 3000L/r_i \rceil \times d_m}$ for the $i$-th pyramidal feature sequence can be formulated as

$$\mathbf{H}_i^{(L)} = \mathrm{TransformerEncoder}(\mathbf{Z}_i^{(L)}), \tag{5}$$

where $\mathrm{TransformerEncoder}(\cdot)$ denotes the encoder component of the transformer. To integrate the output hidden states at various time steps into a single feature vector, we utilized the attention layer [53] [54] as [17]. Foremostly, the output hidden states $\mathbf{H}_i^{(L)}$ were projected into the attentional hidden states $\mathbf{A}_i^{(L)} = \{\mathbf{a}_{i,1}, \mathbf{a}_{i,2}, ..., \mathbf{a}_{i,T_i}\}$, where $T_i$ is $\lceil 3000L/r_i \rceil$, via a single fully-connected layer. Thereafter, the attentional feature vector $\bar{\mathbf{a}}_i$ of the $i$-th pyramidal feature sequence can be formulated via a weighted summation of $\mathbf{a}_{i,t}$ along the temporal dimension:

$$\bar{\mathbf{a}}_i = \sum_{t=1}^{T_i} \alpha_{i,t} \mathbf{a}_{i,t}, \tag{6}$$

where the attention weight $\alpha_{i,t}$ at time step $t$ is derived by

$$\alpha_{i,t} = \frac{\exp(s(\mathbf{a}_{i,t}))}{\sum_t \exp(s(\mathbf{a}_{i,t}))}, \tag{7}$$

where $s(\cdot)$ denotes the attention scoring function that maps the attentional hidden state into a scalar value. Formally, this function is defined as

$$s(\mathbf{a}) = \mathbf{a}\mathbf{W}_{\mathrm{att}}, \tag{8}$$

where $\mathbf{W}_{\mathrm{att}} \in \mathbb{R}^{d_m \times 1}$ represents the trainable parameter.

Upon using the attention feature vector $\bar{\mathbf{a}}$ obtained from Eq. 6, the output logits of the $i$-th pyramidal feature sequence, $\mathbf{o}_i$, can be formulated as follows:

$$\mathbf{o}_i = \bar{\mathbf{a}}_i \mathbf{W}_{\mathrm{a}} + \mathbf{b}_{\mathrm{a}}, \tag{9}$$

where $\mathbf{W}_{\mathrm{a}} \in \mathbb{R}^{d_m \times N_c}$ and $\mathbf{b}_{\mathrm{a}} \in \mathbb{R}^{1 \times N_c}$ denote the trainable weight and bias, respectively. Eventually, the sleep stage $\hat{y}^{(L)}$ of the target epoch was predicted based on the following relation.

$$\hat{y}^{(L)} = \mathrm{argmax}\Big(\sum_{i \in \{3,4,5\}} \mathbf{o}_i\Big), \tag{10}$$

where $\mathrm{argmax}(\cdot)$ is the operation that returns the index of the maximum value.

## 3 SUPERVISED CONTRASTIVE LEARNING

As illustrated in Fig. 2, our learning framework involves two training steps. The first step involves contrastive representation learning (*CRL*) to pretrain the backbone network $f(\cdot)$ via a contrastive learning strategy. In this step, the backbone network $f(\cdot)$ is trained to extract the class discriminative features based on the supervised contrastive loss [43]. The second step involves multiscale temporal context learning (*MTCL*) to learn sequential relations in feature pyramid. For this purpose, we acquire the learned weights of $f(\cdot)$ from *CRL* and freeze them to conserve the class discriminative features. The remaining parts of the network ($g(\cdot)$ and $h(\cdot)$) are trained to learn the multiscale temporal context by predicting the sleep stage of the target epoch.

To prevent overfitting during the training procedures, we performed early stopping in both *CRL* and *MTCL*. Thus, validation was performed to track the validation cost (*i.e.*, validation loss) at every certain training iteration (*i.e.*, validation period, $\psi$), and the training was terminated if the validation loss did not progress more than a certain number of times (*i.e.*, early stopping patience, $\phi$). Early stopping in supervised contrastive learning (*SCL*) results in better pretrained parameters of the backbone network and prevents overfitting at the *MTCL* step. The details of the hyperparameters used in *SCL* are summarized in Section 4.4. Note that different validation periods, $\psi_1$ and $\psi_2$, were used in *CRL* and *MTCL*, respectively. The specifics of the training framework are described in the following sections and Algorithm 1.

### 3.1 Contrastive Representation Learning

In *CRL*, we implemented the training strategy of *SCL* [43] to extract the class discriminative features from a single EEG epoch. As illustrated in Fig. 2-(a), the *CRL* aimed to maximize the similarity between the two projected vectors based on two different views of a single EEG epoch. Simultaneously, the similarity between two projected vectors from different sleep stages was minimized as the optimization of supervised contrastive loss. Thus, we focused on reducing the ambiguous frequency characteristics by extracting distinguishable representations of the sleep stage. Accordingly, a single EEG epoch was initially transformed by two distinct augmentation functions. Thereafter, the encoder network and projection network mapped them into the hypersphere, which produced a latent vector $z$. The details are explained in the following sections: *Augmentation Module*, *Encoder Network*, *Projection Network*, and *Loss Function*, which constitute the major components of *CRL*.

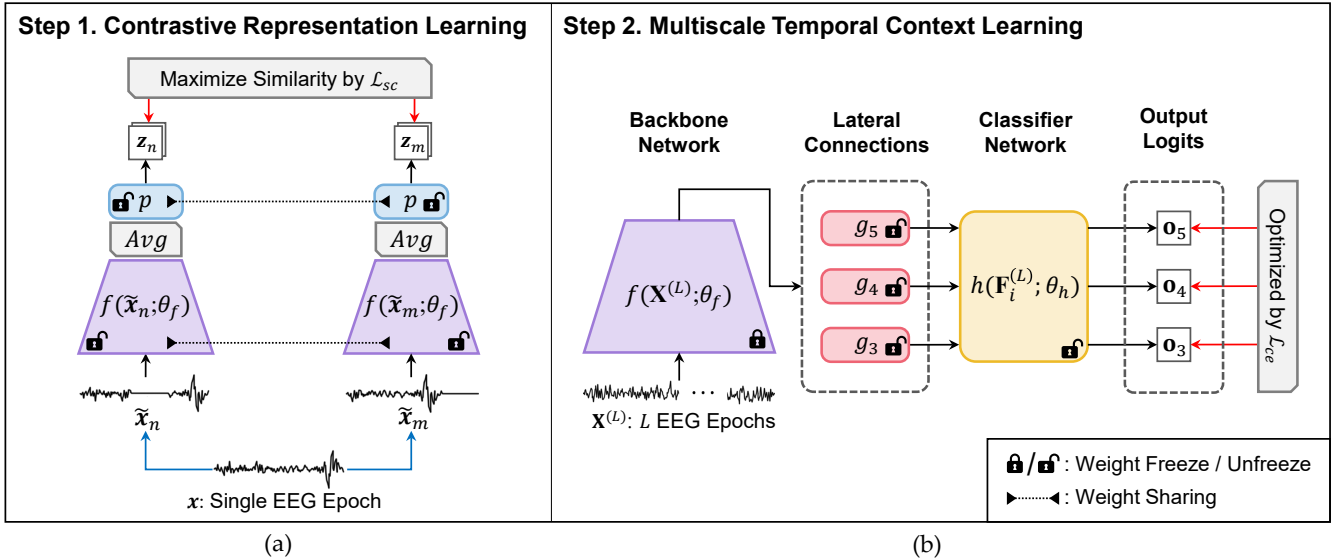

Figure 2: Illustration of the proposed supervised contrastive learning framework. (a) Contrastive representation learning (*CRL*); (b) multiscale temporal context learning (*MTCL*); red arrow is first backward path and blue arrow is $Aug(\cdot)$.

**Data Augmentation Module:** The data augmentation module, $Aug(\cdot)$, transforms a single epoch of EEG signal $x$ into a slightly different but semantically identical signal as follows:

$$\tilde{x} = Aug(x), \quad (11)$$

where $\tilde{x}$ denotes a different perspective on a single EEG epoch $x$. With a given set of randomly sampled data $\{x_p, y_p\}_{p=1,\ldots,N_b}$ (*i.e.*, batch), two different augmentations result in $\{\tilde{x}_q, \tilde{y}_q\}_{q=1,\ldots,2N_b}$, called a multiviewed batch [43] as illustrated in Fig. 2-(a), where $N_b$ is the batch size and $\tilde{y}_{2p-1} = \tilde{y}_{2p} = y_p$. Because the augmentation set is crucial for contrastive learning, we adopted the verified transformation functions from previous studies [23], [33]. As listed in Table 1, we sequentially applied six transformations, namely, *amplitude scale*, *time shift*, *amplitude shift*, *zero-masking*, *additive Gaussian noise*, and *band-stop filter* with the probability of 0.5. In addition, we modified the hyperparameter of the transformation functions by considering the sampling rate and characteristics of the EEG signal.

**Encoder Network:** The encoder network transforms an augmented single-EEG epoch $\tilde{x}$ into a representation vector $r \in \mathbb{R}^{c_5}$. The encoder network contains the sequence of the backbone network $f(\cdot)$ and the global average pooling operation $Avg(\cdot)$. Thus, the backbone network initially transforms an augmented single-EEG epoch $\tilde{x}$ into the feature sequence $\mathbf{F}_5^{(1)}$. Then, the representation vector is obtained from the feature sequence via global average pooling along the temporal dimension. Formally, the representation vector

Table 1: Data Augmentation Pipeline

| Transformation Pipeline | Min | Max | Probability |
|---|---|---|---|
| amplitude scaling | 0.5 | 2 | |
| time shift (samples) | -300 | 300 | |
| amplitude shift ($\mu$V) | -10 | 10 | 0.5 each |
| zero-masking (samples) | 0 | 300 | |
| additive Gaussian noise ($\sigma$) | 0 | 0.2 | |
| band-stop filter (2 Hz width) (lower bound frequency, Hz) | 0.5 | 30.0 | |

is evaluated as $r = Avg(\mathbf{F}_5^{(1)})$.

**Projection Network:** The projection network is vital for mapping the representation vector $r$ into the normalized latent vector $z = \frac{z'}{\|z'\|_2} \in \mathbb{R}^{d_z}$, where $z' = p(r; \theta_p)$, $p(\cdot)$ denotes the projection network, $\theta_p$ represents its trainable parameter, and $d_z$ indicates the dimension of the latent vector. We use a multilayer perceptron [55] with a single hidden layer of size 128 to obtain a latent vector of size $d_z = 128$ as the projection network. For sequential modeling, the $p(\cdot)$ is removed at the *MTCL*.

**Loss Function:** For *CRL*, we adopt the supervised contrastive loss proposed in [43]. The supervised contrastive loss simultaneously maximizes the similarity between positive pairs and promotes the deviations across negative pairs. In this study, samples annotated with the same sleep stage in a multiviewed batch are regarded as positives, and negatives otherwise. Formally, the supervised contrastive loss function can be formulated as

$$\mathcal{L}_{sc} = -\sum_{n=1}^{2N_b} \frac{1}{N_p^{(n)}} \sum_{m=1}^{2N_b} \log \frac{\mathbb{1}_{[n \neq m]} \mathbb{1}_{[\tilde{y}_n = \tilde{y}_m]} \exp(z_n \cdot z_m / \tau)}{\sum_{a=1}^{2N_b} \mathbb{1}_{[n \neq a]} \exp(z_n \cdot z_a / \tau)}, \quad (12)$$

where $N_p^{(n)}$ denotes the number of positives for the $n$-th sample in a multiviewed batch excluding itself, $\mathbb{1}_{[\cdot]}$ denotes the indicator function, $\tilde{y}_n$ denotes the ground truth corresponding to $z_n$, · operation denotes the inner product between two vectors, and $\tau \in \mathbb{R}^+$ denotes a scalar temperature parameter ($\tau = 0.07$ in all present experiments).

### 3.2 Multiscale Temporal Context Learning

As illustrated in Fig. 2-(b), the second step of *SleePyCo* executes $L$ successive EEG epochs $\mathbf{X}^{(L)}$ to analyze both the intra- and inter-epoch temporal contexts ($L = 10$ in this study). The performance obtained considering intra- and inter-epoch temporal contexts [15] is better than that considering only the intra-epoch. However, it is difficult for the network to capture the EEG patterns of the previous epochs, because only the label of the target epoch is provided. To resolve this, we fixed the parameters of the backbone network $f(\cdot)$ to maintain and utilize the class discriminative features learned from *CRL*. By contrast, the remaining parts of the network, which are lateral connections $g(\cdot)$ and classifier network $h(\cdot)$, are learned to score the sleep stage of the target epoch based on the following loss function:

$$\mathcal{L}_{ce} = -\sum_{i \in \{3,4,5\}} \sum_{j=1}^{N_c} y_j^{(L)} \log \left( \frac{\exp(o_{i,j})}{\sum_{k=1}^{N_c} \exp(o_{i,k})} \right), \quad (13)$$

where $y_j^{(L)}$ denotes the $j$-th element of one-hot encoding label, and $o_{i,j}$ represents the $j$-th element of output logits from the $i$-th convolutional stage. This loss function follows the summation of cross entropy over the output logits from each convolutional block.

Because all pyramidal feature sequences share a single classifier network, the classifier network considers feature sequences across a broad scale. Thus, Eq. 13 facilitates the classifier network to analyze the temporal relation between the EEG patterns at different temporal scales and frequency characteristics. Consequently, the classifier network $h(\cdot)$ models the temporal information based on the pyramidal feature sequences $\mathbf{F}_i^{(L)}$ derived from analyzing the intra- and inter-epoch temporal contexts.

## 4 EXPERIMENTS

### 4.1 Datasets

Four public datasets, including PSG recordings and their associated sleep stages, were utilized to assess the performance of *SleePyCo*: Sleep-EDF (2018 version), Montreal Archive of Sleep Studies (MASS), Physionet2018, and Sleep Heart Health Study (SHHS). The number of subjects, utilized EEG channels, evaluation scheme, number of held-out validation subjects, and sample distribution are presented in Table 2. In this study, all EEG signals except the Sleep-EDF dataset were downsampled to 100 Hz. We did not employ preprocessing to EEG signals except for downsampling. For all datasets, we discarded the EEG epochs with annotations that were not related to the sleep stage, such as MOVEMENT class. In addition, we merged N3 and N4 into N3 to consider the five-class problem for datasets annotated with R&K [4].

**Sleep-EDF:** The Sleep-EDF Expanded dataset [44], [45] (2018 version) includes 197 PSG recordings containing EEG, EOG, chin EMG, and event markers. The Sleep-EDF dataset

---

**Algorithmus 1** Training Algorithm

**Input:** $data_{train}$, $data_{val}$, early stopping patience $\phi$, learning rate $\eta$, validation period $\psi_1, \psi_2$, early stopping counter $p$, iteration counter $q$, and trainable parameter $\theta_f, \theta_p, \theta_g, \theta_h$

/* Step1: Contrastive Representation Learning */
1: $p \leftarrow 0, q \leftarrow 0$
2: **while** $p \leq \phi$ **do**
3:    Sample a minibatch $(\mathbf{X}^{(1)}, \boldsymbol{y}^{(1)}) \in data_{train}$
4:    Calculate $\mathcal{L}_{sc}$ as Eq. 12
5:    Update $\theta_f$ and $\theta_p$ w.r.t $\mathcal{L}_{sc}$ by Adam with $\eta$
6:    $q \leftarrow q + 1$
7:    **if** $q \mod \psi_1 = 0$ **then**
8:      Calculate $\mathcal{L}_{sc}$ for $data_{val}$
9:      **if** $\mathcal{L}_{sc} >$ previous $\mathcal{L}_{sc}$ **then**
10:        $p \leftarrow p + 1$
11:      **else**
12:        Store $\theta_f, p \leftarrow 0$
/* Step2: Multiscale Temporal Context Learning */
13: $p \leftarrow 0, q \leftarrow 0$
14: Restore $\theta_f$ of the lowest $\mathcal{L}_{sc}$, then freeze $\theta_f$
15: **while** $p \leq \phi$ **do**
16:    Sample a minibatch $(\mathbf{X}^{(L)}, \boldsymbol{y}^{(L)}) \in data_{train}$
17:    Calculate $\mathcal{L}_{ce}$ as Eq. 13
18:    Update $\theta_g$ and $\theta_h$ w.r.t $\mathcal{L}_{ce}$ by Adam with $\eta$
19:    $q \leftarrow q + 1$
20:    **if** $q \mod \psi_2 = 0$ **then**
21:      Calculate $\mathcal{L}_{ce}$ for $data_{val}$
22:      **if** $\mathcal{L}_{ce} >$ previous $\mathcal{L}_{ce}$ **then**
23:        $p \leftarrow p + 1$
24:      **else**
25:        Store $\theta_g$ and $\theta_h, p \leftarrow 0$
26: **return** trainable parameter $\theta_f, \theta_g, \theta_h$

---

comprises two kinds of studies: SC for 79 healthy Caucasian individuals without sleep disorders and ST for 22 subjects of a study on the effects of Temazepam on sleep. In this study, the SC recordings (subjects aged 25–101 years) were used based on previous studies [7], [17], [21], [25], [27]. According to the R&K rule [4], sleep experts score each half-minute epoch with one of the eight classes {WAKE, REM, N1, N2, N3, N4, MOVEMENT, UNKNOWN}. Owing to the larger size of the class WAKE group compared to others, we included only 30 min of WAKE epochs before and after the sleep period.

**MASS:** The Montreal Archive of Sleep Studies (MASS) dataset [56] contains PSG recordings from 200 individuals (97 males and 103 females). This dataset includes five subsets (SS1, SS2, SS3, SS4, and SS5) that are classified based on the research purpose and data acquisition protocols. The AASM standard [5] (SS1 and SS3 subsets) or the R&K standard [4] (SS2, SS4, and SS5 subsets) was used for the manual annotation. Specifically, the SS1, SS2, and SS4 subsets were annotated with 20-s EEG epochs instead of SS3 and SS5 subsets. Because 30-s EEG samples are used in *SCL*, 5-s segments of signals before and after the EEG epoch were considered for the case of SS1, SS2, and SS4. In *MTCL*, an equal length of EEG signals was used for all subsets (300 s in this study).

Table 2: Experimental settings and dataset statistics

| Dataset | No. of Subjects | EEG Channel | Experimental Setting | | Class Distribution | | | | | |
|---|---|---|---|---|---|---|---|---|---|---|
| | | | Evaluation Scheme | Held-out Validation Set | Wake | N1 | N2 | N3 | REM | Total |
| Sleep-EDF | 79 | Fpz-Cz | 10-fold CV | 7 subjects | 69,824 (35.8%) | 21,522 (10.8%) | 69,132 (34.7%) | 13,039 (6.5%) | 25,835 (13.0%) | 199,352 |
| MASS | 200 | C4-A1 | 20-fold CV | 10 subjects | 31,184 (13.6%) | 19,359 (8.5%) | 107,930 (47.1%) | 30,383 (13.3%) | 40,184 (17.5%) | 229,040 |
| Physio2018 | 994 | C3-A2 | 5-fold CV | 50 subjects | 157,945 (17.7%) | 136,978 (15.4%) | 377,870 (42.3%) | 102,592 (11.5%) | 116,877 (13.1%) | 892,262 |
| SHHS | 5,793 | C4-A1 | Train/Test: 0.7:0.3 | 100 subjects | 1,691,288 (28.8%) | 217,583 (3.7%) | 2,397,460 (40.9%) | 739,403 (12.6%) | 817,473 (13.9%) | 5,863,207 |

**Physio2018:** The Physio2018 dataset is contributed by the Computational Clinical Neurophysiology Laboratory at Massachusetts General Hospital and was applied to detect arousal during sleep in the 2018 Physionet challenge [44], [57]. Owing to the unavailability of annotation for the evaluation set, we used the training set containing PSG recordings for 994 subjects aged 18–90. Thereafter, these recordings were annotated with the AASM rules [5], and we employed only C3–A2 EEG recordings for the single-channel EEG classification.

**SHHS:** The SHHS dataset [58], [59] is a multicenter cohort research that is designed to examine the influence of sleep apnea on cardiovascular diseases. The collection is composed of two rounds of PSG records: Visit 1 (SHHS-1) and Visit 2 (SHHS-2), wherein each record contains two-channel EEGs, two-channel EOGs, a single-channel EMG, a single-channel ECG, and two-channel respiratory inductance plethysmography, as well as the data from location sensors, light sensors, pulse oximeters, and airflow sensors. Each epoch was assigned a value of W, REM, N1, N2, N3, N4, MOVEMENT, and UNKNOWN according to the R&K rule [4]. In this study, the single-channel EEG (C4–A1) was analyzed from 5,793 SHHS-1 recordings.

### 4.2 Backbone Networks for Ablation Study

A direct comparison between the automatic sleep scoring methods would not be justified depending on the experimental settings such as the data processing method and training framework. For fair comparison with other state-of-the-art backbones, we implemented four backbones in the current training framework: *DeepSleepNet* [16], *IITNet* [15], *U-Time* [25], and *XSleepNet* [7]. The backbones were trained and evaluated in the proposed framework on the same datasets (Sleep-EDF, MASS, Physio2018, and SHHS).

**DeepSleepNet Backbone:** The structure of *DeepSleepNet* [16] consists of a dual path CNN for representation learning and two layers of LSTM for sequential learning. To compare the representation power of the backbone network, we considered only the dual-path CNN component of *DeepSleepNet*. The filter width of a single path is smaller for capturing certain EEG patterns, and that of the other path is larger to consider the frequency components from the EEG. To aggregate features from each CNN path, interpolation is performed after the CNN path with larger filters. Thereafter, we concatenated these two features and applied the two convolutional layers following [15]. The output size from the last layer was $16 \times 128$ with a single EEG epoch.

**IITNet Backbone:** *IITNet* [15] uses the modified one-dimensional ResNet-50 for extracting the representation of the raw EEG signal. Similar to the backbone network of *SleePyCo*, the backbone network of *IITNet* forms a five-block structure. However, the backbone network of *IITNet* has a deep architecture (49 convolutional layers), whereas that of *SleePyCo* is shallow (13 convolutional layers). Given a single EEG epoch of size $3000 \times 1$, the backbone network of *IITNet* outputs feature sequences of sizes $1500 \times 16$, $750 \times 64$, $375 \times 64$, $94 \times 128$, and $47 \times 128$ from each convolutional block.

**U-Time Backbone:** *U-Time* [25] is a fully convolutional network for time-series segmentation applied for automatic sleep scoring. Similar to previous fully convolutional networks [60], [61], *U-Time* is the encoder–decoder structure, with the encoder for feature extraction and the decoder for time-series segmentation. Accordingly, we implemented only the encoder component to capture EEG patterns from raw EEG signals. The *U-Time* backbone comprises five convolutional blocks, similar to the proposed backbone network. However, the output from the last convolutional block could not be calculated from the single-epoch EEG because the encoder was designed to analyze 35 epochs of the EEG signals. To solve this problem, we lengthened the temporal dimension of the feature sequences by modifying the filter width of the max-pooling layer between the convolutional blocks from $\{10, 8, 6, 4\}$ to $\{8, 6, 4, 2\}$, respectively. Consequently, the sizes of the feature sequences were $3000 \times 16$, $375 \times 32$, $62 \times 64$, $15 \times 128$, and $7 \times 256$ from each convolutional block for a single EEG epoch.

**XSleepNet Backbone:** *XSleepNet* [7] is a multiviewed model that acquires raw signals and time–frequency images as inputs. Thus, *XSleepNet* includes two types of encoder network: one for the raw signals and the other for the time–frequency images. To compare the raw signal domain, we used the encoder component on raw signals in the ablation study. This encoder was composed of nine one-dimensional convolutional layers with a filter width of 31 and stride length of 2. The sizes of the output feature sequences were $1500 \times 16$, $750 \times 16$, $325 \times 32$, $163 \times 32$, $82 \times 64$, $41 \times 64$, $21 \times 128$, $11 \times 128$, and $6 \times 256$, as obtained from 3000 samples of input EEG epochs.

### 4.3 Model Specifications

The details of the components of *SleePyCo-backbone* are presented in Table 3. In addition, we used the one-dimensional operations of the convolutional layer, batch normalization layer, and max-pooling layer. All convolutional layers in *SleePyCo-backbone* had a filter width of 3, stride size of 1, and padding size of 1 to maintain the temporal dimension in the convolutional block. The max-pooling layers were utilized with a filter width of 5 between each convolutional block to reduce the temporal dimension of the feature sequence. As explained in Section 2.2.2, the lateral connections that follow the backbone network had a filter width of 1. The outputs of lateral connections bear the same feature dimension $d_f$ of 128. For the transformer encoder of the classifier network, the number of heads $N_h$ was 8, number of encoder layers $N_e$ was 6, model dimension of transformer $d_m$ was 128, and the dimension of the feed-forward network $d_{FF}$ was 128.

Table 3: Model Specification of *SleePyCo-backbone*. $T$: temporal dimension, $C$: channel dimension, BN: Batch Normalization and SE: Squeeze and Excitation [48].

| Layer Name | Composition | Output Size [$T \times C$] $L = 1$ | $L = 10$ |
|---|---|---|---|
| Input | - | $3000 \times 1$ | $30000 \times 1$ |
| Conv1 | 3 Conv + BN + PReLU<br>3 Conv + BN + SE + PReLU | $3000 \times 64$ | $30000 \times 64$ |
| Max-pool1 | 5 Max-pool | $600 \times 64$ | $6000 \times 64$ |
| Conv2 | 3 Conv + BN + PReLU<br>3 Conv + BN + SE + PReLU | $600 \times 128$ | $6000 \times 128$ |
| Max-pool2 | 5 Max-pool | $120 \times 128$ | $1200 \times 128$ |
| Conv3 | 3 Conv + BN + PReLU<br>3 Conv + BN + PReLU<br>3 Conv + BN + SE + PReLU | $120 \times 192$ | $1200 \times 192$ |
| Max-pool3 | 5 Max-pool | $24 \times 192$ | $240 \times 192$ |
| Conv4 | 3 conv + BN + PReLU<br>3 Conv + BN + PReLU<br>3 Conv + BN + SE + PReLU | $24 \times 256$ | $240 \times 256$ |
| Max-pool4 | 5 Max-pool | $5 \times 256$ | $48 \times 256$ |
| Conv5 | 3 Conv + BN + PReLU<br>3 Conv + BN + PReLU<br>3 Conv + BN + SE + PReLU | $5 \times 256$ | $48 \times 256$ |

### 4.4 Model Training

In all experiments, the networks were trained using the Adam optimizer [62] with $\eta = 1 \times 10^{-4}$, $\beta_1 = 0.9$, $\beta_2 = 0.999$, and $\epsilon = 1 \times 10^{-8}$. L2-weight regularization was used with a factor of $1 \times 10^{-6}$ to prevent overfitting. Because a large batch size benefits contrastive learning [33], [38], [43], a batch size of 1,024 was employed in the *SCL*, whereas that of 64 was used in *MTCL*. For all datasets except SHHS, validation was conducted to track the validation loss used for early termination at every 50-th and 500-th training iterations (*i.e.*, validation period, $\psi$) in the *SCL* and *MTCL*, respectively. For the larger dataset SHHS, the validation period was 500 and 5000 in *SCL* and *MTCL*, respectively. Early stopping was employed by tracking the validation loss, such that the training was terminated if the validation loss did not decrease for 20 validation steps, (*i.e.*, early stopping patience, $\phi$). At each cross-validation, the model with the lowest validation loss was evaluated on the test set. The networks were trained on NVIDIA GeForce RTX 3090. Python 3.8.5 and PyTorch 1.7.1 [63] were adopted in this study.

### 4.5 Model Evaluation

#### 4.5.1 Evaluation Scheme

To assess the performance of *SleePyCo*, we conducted $k$-fold cross-validation for the Sleep-EDF, MASS, and Physio2018 datasets. Given that the number of subjects in a dataset is $N_s$, the records with $N_s/k$ subjects were used for model evaluation (*i.e.*, test set), and the other records were classified into training and validation data. The selection of subjects for model evaluation was performed over all subjects by sequentially changing on $k$ folds. As listed in Table 2, we utilized $k$ as 10, 20, and 5 for the Sleep-EDF, MASS, and Physio2018 dataset, respectively. The held-out validation set refers to the number of subjects used for the validation set in a fold. For instance, subjects of the MASS dataset were categorized into 180, 10, and 10 recordings for the training, validation, and test set, respectively. Unlike these datasets, the SHHS dataset was randomly divided in a ratio of 0.7 to 0.3 for training and validation, respectively. As performed in [7], we used 100 subjects for the validation set.

#### 4.5.2 Evaluation Criteria

As the evaluation criteria, the overall accuracy (ACC), macro F1-score (MF1), and Cohen's Kappa ($\kappa$) [64] were used for the overall performance measurement and per-class F1-score (F1) was used for class-specific performance measurement. The respective equations for the evaluation criteria are as follows:

$$\text{ACC} = \frac{\text{TP} + \text{TN}}{\text{TP} + \text{FP} + \text{TN} + \text{FN}}, \quad (14)$$

$$\text{MF1} = \frac{1}{N_c} \sum_{j=1}^{N_c} \text{F1}_j = \frac{1}{N_c} \sum_{j=1}^{N_c} \frac{2 \times \text{PR}_j \times \text{RE}_j}{\text{PR}_j + \text{RE}_j}, \quad (15)$$

$$\kappa = \frac{\text{ACC} - \text{P}_e}{1 - \text{P}_e} = 1 - \frac{1 - \text{ACC}}{1 - \text{P}_e}, \quad (16)$$

where TP is true positive, TN is true negative, FP is false positive, FN is false negative, and $\text{F1}_j$, $\text{PR}_j$, and $\text{RE}_j$ are per-class F1-score, per-class precision, and per-class recall of the $j$-th class, respectively. In Eq. 16, $\text{P}_e$ represents the hypothetical probability of chance agreement. Typically, precision (PR) and recall (RE) can be defined as follows:

$$\text{PR} = \frac{\text{TP}}{\text{TP} + \text{FP}}, \quad (17) \qquad \text{RE} = \frac{\text{TP}}{\text{TP} + \text{FN}}. \quad (18)$$

ACC is the intuitive performance indicator that is generally considered in several classification tasks. However, the F1-score indicating the harmonic mean of precision and recall is more valuable in class-imbalanced tasks such as sleep-stage classification. In addition, the F1-score per class indicates the class-specific performance of the F1-score by calculating Eq. 15 without averaging. Cohen's Kappa $\kappa$ indicates the agreement by chance for imbalanced proportions of various classes with a maximum value of 1.0 for ideal classification.

# 5 RESULTS AND DISCUSSION

## 5.1 Performance Comparison with State-of-the-Art (SOTA) Frameworks

The performances of *SleePyCo* and SOTA frameworks are presented in Table 4 according to the datasets, system name, input types for the deep learning models, epoch length ($L$) simultaneously given for automatic sleep scoring, and number of subjects considered in the study. Fig. 3 illustrates the confusion matrices of *SleePyCo* on the Sleep-EDF, MASS, Physio2018, and SHHS datasets. For all datasets, *SleePyCo* achieved state-of-the art performance compared with other models based on single-channel EEG. Quantitatively, *SleePyCo* delivered the best performance in terms of overall accuracy, MF1, and $\kappa$: 84.6%, 79.0%, 0.787 for Sleep-EDF, 86.8%, 82.5%, 0.811 for MASS, 80.9%, 78.9%, 0.737% for Physio2018, and 87.9%, 80.7%, 0.830 for SHHS, respectively. The performance differences between *SleePyCo* and the SOTA frameworks were +0.6%p, +1.1%p, +0.009 for Sleep-EDF, +1.6%p, +1.9%p, +0.023 for MASS, +0.6%p, +0.3%, +0.005 for Physio2018, and +0.2%p, +0.0%p, +0.002 for SHHS in overall accuracy, MF1, and $\kappa$, respectively. The proposed model achieved SOTA performance with the introduction of the feature pyramid and supervised contrastive learning. The network, particularly the classifier network, could learn the feature sequences with various temporal and frequency scales. Furthermore, contrastive learning enables the backbone network to extract class discriminative features, which reduces the ambiguous EEG characteristics of sleep stages.

The major advantages of *SleePyCo* over other SOTA frameworks include *performance, use of single-channel EEG, and shorter input EEG epochs*, as indicated in Table 4. We achieved remarkable performance by considering only raw single-channel EEG signals as input without adopting any preprocessing or hand-crafted features. By contrast, existing SOTA frameworks utilize both raw signals and time–frequency images (*i.e.*, spectrogram) [7], [27]. Because the performance of automatic sleep scoring relies on several factors [7], the superiority of the raw EEG signal in comparison to the spectrogram could not be verified. However, the proposed model demonstrated SOTA performance by utilizing raw signals with no information loss compared to the time–frequency image. Moreover, we utilized a smaller number of input epochs ($L = 10$) than that in existing SOTA methods, which consider more than 15 EEG epochs. Although the input EEG epochs can be expanded to more than 10 epochs, we found via internal experiment that more than 10 input epochs did not produce any significant performance improvement. In addition, *SleePyCo* is suitable for real-time sleep scoring because it takes the target epoch and its previous epochs as input. This study can be expanded to classify other types of time-series data, such as sound and biosignals, to exploit the advantages of multiscale representation and supervised contrastive learning.

## 5.2 Ablation Studies

To examine the effectiveness of *SleePyCo*, we conducted ablation studies based on the following three aspects: backbone network, feature pyramid (*FP*), and supervised contrastive learning (*SCL*), which are explained in Sections 5.2.1, 5.2.2, and 5.2.3, respectively. The results of the ablation studies are presented in Tables 5 and 6. In Section 5.2.1, we discuss the performance verification for *SleePyCo-backbone* by replacing it with other SOTA backbones. In Sections 5.2.2 and 5.2.3, we demonstrate the effectiveness of *FP* and *SCL* by eliminating the components. Notably, *SleePyCo* that employs only $\mathbf{F}_5^{(L)}$, which was simultaneously trained from scratch, was set as the baseline, denoted as BL. Without *SCL*, the entire network of *SleePyCo* was trained under the same *MTCL* condition.

### 5.2.1 Performance of SleePyCo-backbone

The performances of *SleePyCo-backbone* and SOTA backbones on Sleep-EDF, MASS, Physio2018, and SHHS are compared in Table 5. For fair comparison to examine the performance of *SleePyCo-backbone*, regardless of the feature pyramid, we designed two experimental settings: single-scale and multiscale. In the single-scale setting, we adopted only the feature sequence from the final convolutional layer in *MTCL* to yield the single-scale representation. For multiscale settings, the feature pyramid was utilized by employing feature sequences with three distinct levels. The *DeepSleepNet*, *TinySleepNet*, and *IITNet* backbones were considered only in the single-scale settings owing to their large memory footprint and structural limitation.

In the single-scale settings, the proposed *SleePyCo-backbone* without feature pyramid (*i.e.*, w/o FP) displayed competitive or superior performance compared to the SOTA backbones. The performance differences between *SleePyCo-backbone* w/o FP and the best results of the single-scale backbones in terms of accuracy, MF1, and $\kappa$ were +0.3%p, +0.2%p, +0.005 for Sleep-EDF, +0.0%p, +0.5%p, +0.000 for MASS, +0.2%p, +0.5%p, +0.004 for Physio2018, and −0.3%p, −1.0%p, −0.004 for SHHS, respectively. With the application of the feature pyramid, the performance of the proposed model was superior to that of the *U-Time* and *XSleepNet* backbones. The variations in overall accuracy, MF1, and $\kappa$ between *SleePyCo-backbone* and other SOTA backbones were +0.2%p, +0.2%p, +0.004 for Sleep-EDF, +0.2%p, +0.5%p, +0.004 for MASS, +0.5%p, +0.6%p, +0.006 for Physio2018, and +0.1%p, +0.9%p, +0.002 for SHHS, respectively.

The results of the extensive experiments revealed the superior performance of the proposed *SleePyCo-backbone* compared to that of other network architectures, regardless of the feature pyramid. In addition, the feature pyramid tended to improve the overall sleep scoring performance in the *U-Time* and *XSleepNet* backbones. These results imply that the feature pyramid improves the sleep scoring performance by imparting richer features in *SleePyCo-backbone* as well as other architectures.

### 5.2.2 Influence of Feature Pyramid

Table 6 shows the ablation study results for *FP* and *SCL* on the Sleep-EDF dataset. The results indicate that the feature pyramid improves the sleep scoring performance, regardless of the *SCL*. Upon adding the feature pyramid to BL, the overall performances were enhanced by 0.3%p, 0.4%p, and 0.005 in ACC, MF1, and $\kappa$, respectively. In the case of application to *SCL*, the feature pyramid enhanced

Table 4: Performance comparison between *SleePyCo* and state-of-the-art (SOTA) methods for automatic sleep scoring via deep learning; bold and underline indicate the best and second best, respectively. Furthermore, the results indicated by [†] were not directly comparable because they employed a subset distinct from the corresponding dataset. RS and SP denote Raw Signal and SPectrogram, respectively.

| | Method | | | | Overall Metrics | | | Per-class F1 Score | | | | |
|---|---|---|---|---|---|---|---|---|---|---|---|---|
| Dataset | System | Input | Epoch Length | Subjects | ACC | MF1 | $\kappa$ | W | N1 | N2 | N3 | REM |
| Sleep-EDF | **SleePyCo (Ours)** | RS | **10 (9 past)** | 79 | **84.6** | **79.0** | **0.787** | **93.5** | <u>50.4</u> | **86.5** | <u>80.5</u> | **84.2** |
| Sleep-EDF | XSleepNet [7] | RS + SP | 20 | 79 | <u>84.0</u> | <u>77.9</u> | <u>0.778</u> | <u>93.3</u> | 49.9 | <u>86.0</u> | 78.7 | <u>81.8</u> |
| Sleep-EDF | Korkalainen *et al.* [24] | RS | 100 | 79 | 83.7 | - | 0.77 | - | - | - | - | - |
| Sleep-EDF | TinySleepNet [23] | RS | <u>15</u> | 79 | 83.1 | 78.1 | 0.77 | 92.8 | **51.0** | 85.3 | **81.1** | 80.3 |
| Sleep-EDF | SeqSleepNet [17] | SP | 20 | 79 | 82.6 | 76.4 | 0.760 | - | - | - | - | - |
| Sleep-EDF | SleepTransformer [27] | SP | 21 | 79 | 81.4 | 74.3 | 0.743 | 91.7 | 40.4 | 84.3 | 77.9 | 77.2 |
| Sleep-EDF | U-Time [25] | RS | 35 | 79 | 81.3 | 76.3 | 0.745 | 92.0 | **51.0** | 83.5 | 74.6 | 80.2 |
| Sleep-EDF | SleepEEGNet [21] | RS | **10** | 79 | 80.0 | 73.6 | 0.73 | 91.7 | 44.1 | 82.5 | 73.5 | 76.1 |
| MASS | **SleePyCo (Ours)** | RS | **10 (9 past)** | 200 | **86.8** | **82.5** | **0.811** | **89.2** | **60.1** | **90.4** | **83.8** | **89.1** |
| MASS | XSleepNet [7] | RS + SP | <u>20</u> | 200 | <u>85.2</u> | <u>80.6</u> | <u>0.788</u> | - | - | - | - | - |
| MASS | SeqSleepNet [17] | SP | 20 | 200 | 84.5 | 79.8 | 0.778 | - | - | - | - | - |
| MASS[†] | Sun *et al.* [22] | RS + SP | 25 | 147 | 86.1 | 79.6 | 0.795 | 85.1 | 50.2 | 89.8 | 84.0 | 89.0 |
| MASS[†] | Qu *et al.* [28] | RS | 30 | 62 | 86.5 | 81.0 | 0.799 | 87.2 | 52.8 | 91.5 | 87.0 | 86.6 |
| MASS[†] | IITNet [15] | RS | **10 (9 past)** | 62 | 86.3 | 80.5 | 0.794 | 85.4 | 54.1 | 91.3 | 86.8 | 84.8 |
| MASS[†] | DeepSleepNet [16] | RS | whole night epochs | 62 | 86.2 | 81.7 | 0.800 | 87.3 | 59.8 | 90.3 | 81.5 | 89.3 |
| Physio2018 | **SleePyCo (Ours)** | RS | **10 (9 past)** | 994 | **80.9** | **78.9** | **0.737** | **84.2** | **59.3** | **85.3** | **79.4** | **86.3** |
| Physio2018 | XSleepNet [7] | RS + SP | <u>20</u> | 994 | <u>80.3</u> | <u>78.6</u> | <u>0.732</u> | - | - | - | - | - |
| Physio2018 | SeqSleepNet [17] | SP | 20 | 994 | 79.4 | 77.6 | 0.719 | - | - | - | - | - |
| Physio2018 | U-Time [25] | RS | 35 | 994 | 78.8 | 77.4 | 0.714 | <u>82.5</u> | <u>59.0</u> | <u>83.1</u> | <u>79.0</u> | <u>83.5</u> |
| SHHS | **SleePyCo (Ours)** | RS | <u>10 (9 past)</u> | 5,793 | **87.9** | **80.7** | **0.830** | **92.6** | <u>49.2</u> | **88.5** | 84.5 | **88.6** |
| SHHS | SleepTransformer [27] | SP | 21 | 5,791 | <u>87.7</u> | <u>80.1</u> | <u>0.828</u> | <u>92.2</u> | 46.1 | <u>88.3</u> | <u>85.2</u> | **88.6** |
| SHHS | XSleepNet [7] | RS + SP | 20 | 5,791 | 87.6 | **80.7** | 0.826 | 92.0 | <u>49.9</u> | <u>88.3</u> | 85.0 | <u>88.2</u> |
| SHHS | Sors *et al.* [14] | RS | **4 (2 past, 1 future)** | 5,728 | 86.8 | 78.5 | 0.815 | 91.4 | 42.7 | 88.0 | 84.9 | 85.4 |
| SHHS | IITNet [15] | RS | <u>10 (9 past)</u> | 5,791 | 86.7 | 79.8 | 0.812 | 90.1 | 48.1 | 88.4 | **85.2** | 87.2 |
| SHHS | SeqSleepNet [17] | SP | 20 | 5,791 | 86.5 | 78.5 | 0.81 | - | - | - | - | - |

| Dataset | Sleep-EDF | | | | | MASS | | | | | Physio2018 | | | | | SHHS | | | | |
|---|---|---|---|---|---|---|---|---|---|---|---|---|---|---|---|---|---|---|---|---|
| AC \ PC | W | N1 | N2 | N3 | REM | W | N1 | N2 | N3 | REM | W | N1 | N2 | N3 | REM | W | N1 | N2 | N3 | REM |
| W | 63,640 (93.0%) | 3,739 (5.5%) | 547 (0.8%) | 29 (0.0%) | 492 (0.7%) | 26,022 (89.1%) | 1,960 (6.7%) | 640 (2.2%) | 37 (0.1%) | 531 (1.8%) | 132,475 (87.3%) | 14,656 (9.7%) | 3,866 (2.5%) | 156 (0.1%) | 650 (0.4%) | 461,447 (93.5%) | 6,500 (1.3%) | 18,513 (3.8%) | 1,586 (0.3%) | 5,367 (1.1%) |
| N1 | 3,281 (15.2%) | 9,881 (45.9%) | 6,603 (30.7%) | 50 (0.2%) | 1,707 (7.9%) | 1,865 (9.7%) | 10,532 (54.8%) | 3,925 (20.4%) | 12 (0.1%) | 2,875 (15.0%) | 22,437 (16.7%) | 73,624 (54.6%) | 30,750 (22.8%) | 139 (0.1%) | 7,797 (5.8%) | 15,077 (23.2%) | 28,570 (43.9%) | 14,861 (22.9%) | 18 (0.0%) | 6,486 (10.0%) |
| N2 | 402 (0.6%) | 2,824 (4.1%) | 62,247 (90.0%) | 1,689 (2.4%) | 1,970 (2.8%) | 697 (0.6%) | 2,136 (2.0%) | 98,472 (91.3%) | 4,399 (4.1%) | 2,145 (2.0%) | 5,461 (1.4%) | 19,007 (5.0%) | 329,927 (87.4%) | 16,186 (4.3%) | 6,760 (1.8%) | 19,273 (2.7%) | 12,433 (1.7%) | 636,895 (88.7%) | 29,457 (4.1%) | 19,587 (2.7%) |
| N3 | 38 (0.3%) | 22 (0.2%) | 2,982 (22.9%) | 9,984 (76.6%) | 13 (0.1%) | 69 (0.2%) | 10 (0.0%) | 5,176 (17.0%) | 25,121 (82.7%) | 7 (0.0%) | 336 (0.3%) | 110 (0.1%) | 23,617 (23.0%) | 78,430 (76.5%) | 94 (0.1%) | 899 (0.4%) | 5 (0.0%) | 36,061 (16.1%) | 186,981 (83.4%) | 240 (0.1%) |
| REM | 271 (1.0%) | 1,247 (4.8%) | 2,477 (9.6%) | 10 (0.0%) | 21,830 (84.5%) | 475 (1.2%) | 1,186 (3.0%) | 1,740 (4.3%) | 4 (0.0%) | 36,779 (91.5%) | 2,138 (1.8%) | 6,084 (5.2%) | 8,307 (7.1%) | 101 (0.1%) | 100,208 (85.8%) | 6,534 (2.7%) | 3,521 (1.4%) | 14,650 (6.0%) | 110 (0.0%) | 218,917 (89.8%) |

Figure 3: Confusion matrix of *SleePyCo* for Sleep-EDF, MASS, Physio2018, and SHHS dataset. The values in parentheses indicate per-class recall. AC and PC denote Actual Class and Predicted Class, respectively

Table 5: Performance comparison of *SleePyCo-backbone* and SOTA backbones on experimental datasets; bold and underline indicate the first and second highest, respectively. FP and SCL denote Feature Pyramid and Supervised Contrastive Learning, respectively.

| Backbone | Method | Sleep-EDF | | | MASS | | | Physio2018 | | | SHHS | | |
| --- | --- | --- | --- | --- | --- | --- | --- | --- | --- | --- | --- | --- | --- |
| | | ACC | MF1 | $\kappa$ | ACC | MF1 | $\kappa$ | ACC | MF1 | $\kappa$ | ACC | MF1 | $\kappa$ |
| DeepSleepNet | SCL | <u>83.8</u> | <u>78.2</u> | <u>0.775</u> | <u>86.2</u> | 81.4 | 0.801 | 79.6 | 77.3 | 0.721 | 87.1 | 79.4 | 0.818 |
| IITNet | SCL | 83.5 | 77.8 | 0.771 | 86.1 | 81.4 | 0.801 | 79.9 | 77.6 | 0.724 | **87.5** | <u>79.3</u> | **0.824** |
| U-Time | SCL | 83.6 | 78.1 | 0.773 | **86.4** | <u>81.6</u> | <u>0.804</u> | <u>80.1</u> | <u>77.8</u> | <u>0.726</u> | 87.1 | 79.1 | 0.817 |
| XSleepNet | SCL | 83.4 | 77.2 | 0.769 | 86.0 | 81.3 | 0.799 | 79.7 | 77.6 | 0.722 | <u>87.4</u> | 79.5 | <u>0.821</u> |
| **SleePyCo (Ours)** | SCL | **84.1** | **78.4** | **0.780** | **86.4** | **82.1** | **0.804** | **80.3** | **78.3** | **0.730** | 87.2 | 78.5 | 0.820 |
| U-Time | FP + SCL | <u>84.4</u> | <u>78.8</u> | <u>0.783</u> | <u>86.6</u> | <u>82.0</u> | <u>0.807</u> | <u>80.4</u> | <u>78.3</u> | <u>0.731</u> | <u>87.8</u> | 79.5 | <u>0.828</u> |
| XSleepNet | FP + SCL | 83.5 | 77.5 | 0.771 | 86.2 | 81.8 | 0.803 | 80.2 | 78.0 | 0.728 | 87.7 | <u>79.8</u> | 0.826 |
| **SleePyCo (Ours)** | FP + SCL | **84.6** | **79.0** | **0.787** | **86.8** | **82.5** | **0.811** | **80.9** | **78.9** | **0.737** | **87.9** | **80.7** | **0.830** |

Table 6: Ablation study on Sleep-EDF; bold indicates the highest. BL, FP, and SCL indicate BaseLine, Feature Pyramid, and Supervised Contrastive Learning, respectively.

| Method | Overall Metrics | | | Per-class F1 Score | | | | |
| --- | --- | --- | --- | --- | --- | --- | --- | --- |
| | ACC | MF1 | $\kappa$ | W | N1 | N2 | N3 | REM |
| BL | 83.2 | 77.3 | 0.767 | 93.2 | 46.6 | 85.1 | 79.9 | 81.6 |
| BL + FP | 83.5 | 77.7 | 0.772 | 93.2 | 47.9 | 85.1 | 79.9 | 82.3 |
| BL + SCL | 84.1 | 78.4 | 0.780 | 93.2 | 49.3 | 86.1 | 79.7 | 83.5 |
| **BL + FP + SCL (Ours)** | **84.6** | **79.0** | **0.787** | **93.5** | **50.4** | **86.5** | **80.5** | **84.2** |

the sleep scoring performance by 0.5%p, 0.6%p, and 0.007 in ACC, MF1, and $\kappa$, respectively. Because the proposed model predicts the sleep stage with the summation of logits from each convolutional stage, the feature pyramid has the ensemble effect that enhances the performance based on predictions from various models. As reported in the literature [42], the feature pyramid provides richer information than the single-scale representation, which result in overall performance improvement.

In terms of per-class performance, the F1 scores of entire classes increased when feature pyramid was employed. On average, feature pyramid enhanced the F1 scores by 0.1%p for W, 1.2%p for N1, 0.2%p for N2, 0.4%p, and 0.7%p for REM. The performance improvement for N1, N3, and REM was greater than that for the other sleep stages. This result indicates that low- and mid-level features contribute to the network classification performance for N1, N3, and REM. According to the AASM [5] rule, which is briefly described in Table 8, the fundamental scoring rationales of N1 and REM in EEG are low amplitude, mixed frequency (LAMF) activity as a common feature, and vertex sharp waves and sawtooth waves as a distinguishable feature, respectively. These rationales have a relatively mid-range frequency (4–7 Hz for LAMF activity, 5–14 Hz for vertex sharp waves, and 2–6 Hz for sawtooth waves) compared to the characteristics of the other sleep stages. Moreover, N3 is principally scored by the existence of slow wave activity that has a relatively low-range frequency between 0.5–2 Hz.

To determine the relationship between the feature pyramid and per-class performance, we evaluated the per-class F1 of each stage by separating the output logits from each stage, as described in Table 7. Consequently, the F1 of W and N2, which is scored according to mid- and high-frequency characteristics (W: alpha (8–13 Hz) and beta rhythm (13 – 30 Hz); N2: K complex (8–16 Hz) and sleep spindles (12–14 Hz); REM: mixed frequency (2–14 Hz)), was the highest at stage index 5. At the sleep stage of N1, where the mid-frequency range forms the dominant frequency component (theta wave with 4–7 Hz), the highest F1 occurs at stage index 4. In the case of N3, F1 is significantly high at stage index 3 because N1 exhibits low-frequency characteristics (slow wave activity with 0.5–2 Hz). The results indicate that the specific frequency components were intensively considered according to the feature level. These results demonstrate that the feature pyramid provides more information to enable the consideration of the AASM rules by the network instead of single-scale baselines. Furthermore, the feature type extracted from the CNN varies with the feature level in automatic sleep scoring as well as computer vision [46], [65].

Table 7: Per-class F1 score evaluated on each stage index $i \in \{3, 4, 5\}$; bold indicates the highest except values in the last row.

| Stage Index | Overall Metrics | | | Per-class F1 Score | | | | |
| --- | --- | --- | --- | --- | --- | --- | --- | --- |
| | ACC | MF1 | $\kappa$ | W | N1 | N2 | N3 | REM |
| 3 | 83.9 | 78.1 | 0.776 | 93.2 | 48.7 | 85.9 | **80.3** | 82.5 |
| 4 | **84.4** | **78.7** | **0.783** | 93.4 | **50.4** | 86.2 | 80.0 | **83.7** |
| 5 | **84.4** | 78.5 | **0.783** | **93.5** | 49.7 | **86.3** | 79.4 | **83.7** |
| 3, 4, 5 | 84.6 | 79.0 | 0.787 | 93.5 | 50.4 | 86.5 | 80.5 | 84.2 |

### 5.2.3 Effect of Supervised Contrastive Learning

As observed in Table 6 by comparing BL with BL+SCL and BL+FP with BL+FP+SCL, the overall metrics were improved by *SCL*. When the baseline model w/o and w/ feature pyramid were trained on *SCL*, the overall performances in ACC, MF1, and $\kappa$ were enhanced by 1.0%p, 1.2%p, and 0.014, respectively. In particular, the F1 scores of N1, N2, and REM were significantly improved by 2.6%p, 1.2%p, and 1.9%p, respectively, compared to the improvements of the other classes; 0.3%p for W, 0.6%p for N3 with arithmetic mean. Following the AASM rule, the categorization of sleep stages using only EEG is confusing owing to their similar frequency activities as described in Table 8. Specifically, N1, N2, and REM share a similar EEG characteristic, namely, LAMF activity. Thus, ambiguous EEG characteristics between the sleep stages is a primary factor that affects the sleep scoring performance on EEG signals. Based on these facts and experimental results, the proposed supervised contrastive learning method enables the network to extract the discriminative features between sleep stages more effectively, especially for N1, N2, and REM, than the vanilla-supervised strategy. Primarily, these factors contribute toward the improvement of the overall performance.

Table 8: EEG characteristics of each sleep stage according to the AASM rule [5]; bold indicates the fundamental rationale that scores the corresponding sleep stage.

| Sleep Stage | EEG Characteristics |
|---|---|
| Wake | • **More than 50% of alpha rhythm (8–13 Hz)**<br>• Beta rhythm (13–30 Hz) |
| N1 | • **Vertex sharp waves (5–14 Hz)**<br>• **Low amplitude, mixed frequency activity (4–7 Hz)**<br>• Less than 50% of alpha rhythm (8–13 Hz) |
| N2 | • **K complex (8–16 Hz)**<br>• **Sleep spindle (12–14 Hz)**<br>• Low amplitude, mixed frequency activity (4–7 Hz)<br>• Less than 50% of alpha rhythm (8–13 Hz) |
| N3 | • **More than 20% of slow wave activity (0.5–2 Hz)**<br>• Sleep spindle (12–14 Hz) |
| REM | • **Sawtooth waves (2–6 Hz)**<br>• **Low amplitude, mixed frequency activity (4–7 Hz)**<br>• Alpha rhythm (8–13 Hz)<br>• K complex (8–16 Hz)<br>• Sleep spindle (12–14 Hz) |

In summary, the introduction of the feature pyramid and supervised contrastive learning enhanced the overall performance of the baseline by a considerable margin of 1.4%p, 1.7%p, and 0.02 in ACC, MF1, and $\kappa$, respectively. For the per-class performance, the *F1 scores of N1 and REM were significantly improved by 3.8%p and 2.6%p for REM, respectively*, compared to the other classes (0.3%p for W, 1.4%p for N2, and 0.6%p for N3). These results manifest the synergy acting between the proposed model and supervised contrastive learning to derive an improved representation from the raw EEG signals. The feature pyramid provides richer information by considering various frequency characteristics and temporal scales, and the proposed *SCL* allows the backbone network to extract class-discriminative features that are advantageous for the scoring target epoch.

### 5.3 Limitations and Future Work

The limitations of this study include the chronic problem affecting the transformer, reported in [66], where the computational burden drastically increases for larger feature sequences. The resolution of this issue forms the crux of future work, as the current study was not aimed at designing a computationally efficient model. In addition, we aim to achieve more accurate sleep scoring via supervised contrastive learning for multi-channel PSG signals

## 6 Conclusion

We presented *SleePyCo*, which incorporates a feature pyramid and supervised contrastive learning for accurate automatic sleep scoring. Inspired by the evidence that EEG patterns reflecting the sleep stage can be observed over various temporal and frequency scales, we proposed a deep learning model based on the feature pyramid. The proposed supervised contrastive learning framework reduces the ambiguities between sleep stages by extracting the less-discriminative features. In ablation studies, *SleePyCo-backbone* outperformed SOTA backbones and the feature pyramid and supervised contrastive learning exhibited a synergistic effect in classifying the raw EEG signals. Therefore, the feature pyramid improved the overall performance by considering various temporal and frequency scales of the feature sequences. According to the feature levels and the analysis of the frequency characteristics of the sleep stage, the proposed model exhibited a per-class performance effect. The supervised contrastive learning contributes toward overall performance improvement, which is attributed to the significantly enhanced prediction of N1 and REM by reducing the ambiguity between sleep stages. The comparative analysis demonstrated the SOTA performance of *SleePyCo* on four public datasets of varying sizes: Sleep-EDF, MASS, Physio2018, and SHHS. Furthermore, *SleePyCo* can be expanded to categorize other types of time-series data, which should focus on various temporal and frequency scales and similar frequency characteristics between classes.

### Acknowledgments

This research was supported by a grant from the Institute of Information and Communications Technology Planning and Evaluation (IITP) funded by the Korean government (MSIT) (No. 2020-0-00857, Development of cloud robot intelligence augmentation, sharing and framework technology to integrate and enhance the intelligence of multiple robots). Furthermore, this research was partially supported by the Korea Institute of Energy Technology Evaluation and Planning (KETEP) grant funded by the Korean government (MOTIE) (No. 20202910100030).

### References

[1] K. Wulff, S. Gatti, J. G. Wettstein, and R. G. Foster, "Sleep and circadian rhythm disruption in psychiatric and neurodegenerative disease," *Nature Reviews Neuroscience*, vol. 11, no. 8, pp. 589–599, 2010.

[2] M. Torabi-Nami, S. Mehrabi, A. Borhani-Haghighi, and S. Derman, "Withstanding the obstructive sleep apnea syndrome at the expense of arousal instability, altered cerebral autoregulation and neurocognitive decline," *Journal of integrative neuroscience*, vol. 14, no. 02, pp. 169–193, 2015.

[3] C. Berthomier, X. Drouot, M. Herman-Stoïca, P. Berthomier, J. Prado, D. Bokar-Thire, O. Benoit, J. Mattout, and M.-P. d'Ortho, "Automatic analysis of single-channel sleep eeg: validation in healthy individuals," *Sleep*, vol. 30, no. 11, pp. 1587–1595, 2007.

[4] A. Rechtschaffen, "A manual for standardized terminology, techniques and scoring system for sleep stages in human subjects," *Brain information service*, 1968.

[5] R. B. Berry, R. Brooks, C. E. Gamaldo, S. M. Harding, C. Marcus, B. V. Vaughn *et al.*, "The aasm manual for the scoring of sleep and associated events," *Rules, Terminology and Technical Specifications, Darien, Illinois, American Academy of Sleep Medicine*, vol. 176, p. 2012, 2012.

[6] A. Malhotra, M. Younes, S. T. Kuna, R. Benca, C. A. Kushida, J. Walsh, A. Hanlon, B. Staley, A. I. Pack, and G. W. Pien, "Performance of an automated polysomnography scoring system versus computer-assisted manual scoring," *Sleep*, vol. 36, no. 4, pp. 573–582, 2013.

[7] H. Phan, O. Y. Chén, M. C. Tran, P. Koch, A. Mertins, and M. De Vos, "Xsleepnet: Multi-view sequential model for automatic sleep staging," *IEEE Transactions on Pattern Analysis and Machine Intelligence*, 2021.

[8] J. B. Stephansen, A. N. Olesen, M. Olsen, A. Ambati, E. B. Leary, H. E. Moore, O. Carrillo, L. Lin, F. Han, H. Yan et al., "Neural network analysis of sleep stages enables efficient diagnosis of narcolepsy," *Nature communications*, vol. 9, no. 1, pp. 1–15, 2018.

[9] H. Phan, F. Andreotti, N. Cooray, O. Y. Chén, and M. De Vos, "Dnn filter bank improves 1-max pooling cnn for single-channel eeg automatic sleep stage classification," in *2018 40th annual international conference of the IEEE engineering in medicine and biology society (EMBC)*. IEEE, 2018, pp. 453–456.

[10] ——, "Automatic sleep stage classification using single-channel eeg: Learning sequential features with attention-based recurrent neural networks," in *2018 40th annual international conference of the IEEE engineering in medicine and biology society (EMBC)*. IEEE, 2018, pp. 1452–1455.

[11] ——, "Joint classification and prediction cnn framework for automatic sleep stage classification," *IEEE Transactions on Biomedical Engineering*, vol. 66, no. 5, pp. 1285–1296, 2018.

[12] O. Tsinalis, P. M. Matthews, Y. Guo, and S. Zafeiriou, "Automatic sleep stage scoring with single-channel eeg using convolutional neural networks," *arXiv preprint arXiv:1610.01683*, 2016.

[13] S. Chambon, M. N. Galtier, P. J. Arnal, G. Wainrib, and A. Gramfort, "A deep learning architecture for temporal sleep stage classification using multivariate and multimodal time series," *IEEE Transactions on Neural Systems and Rehabilitation Engineering*, vol. 26, no. 4, pp. 758–769, 2018.

[14] A. Sors, S. Bonnet, S. Mirek, L. Vercueil, and J.-F. Payen, "A convolutional neural network for sleep stage scoring from raw single-channel eeg," *Biomedical Signal Processing and Control*, vol. 42, pp. 107–114, 2018.

[15] H. Seo, S. Back, S. Lee, D. Park, T. Kim, and K. Lee, "Intra- and inter-epoch temporal context network (iitnet) using sub-epoch features for automatic sleep scoring on raw single-channel eeg," *Biomedical Signal Processing and Control*, vol. 61, p. 102037, 2020.

[16] A. Supratak, H. Dong, C. Wu, and Y. Guo, "Deepsleepnet: A model for automatic sleep stage scoring based on raw single-channel eeg," *IEEE Transactions on Neural Systems and Rehabilitation Engineering*, vol. 25, no. 11, pp. 1998–2008, 2017.

[17] H. Phan, F. Andreotti, N. Cooray, O. Y. Chén, and M. De Vos, "Seqsleepnet: end-to-end hierarchical recurrent neural network for sequence-to-sequence automatic sleep staging," *IEEE Transactions on Neural Systems and Rehabilitation Engineering*, vol. 27, no. 3, pp. 400–410, 2019.

[18] A. Vilamala, K. H. Madsen, and L. K. Hansen, "Deep convolutional neural networks for interpretable analysis of eeg sleep stage scoring," in *2017 IEEE 27th international workshop on machine learning for signal processing (MLSP)*. IEEE, 2017, pp. 1–6.

[19] F. Andreotti, H. Phan, N. Cooray, C. Lo, M. T. Hu, and M. De Vos, "Multichannel sleep stage classification and transfer learning using convolutional neural networks," in *2018 40th annual international conference of the IEEE Engineering in Medicine and Biology Society (EMBC)*. IEEE, 2018, pp. 171–174.

[20] M. Längkvist, L. Karlsson, and A. Loutfi, "Sleep stage classification using unsupervised feature learning," *Advances in Artificial Neural Systems*, vol. 2012, 2012.

[21] S. Mousavi, F. Afghah, and U. R. Acharya, "Sleepeegnet: Automated sleep stage scoring with sequence to sequence deep learning approach," *PloS one*, vol. 14, no. 5, p. e0216456, 2019.

[22] C. Sun, C. Chen, W. Li, J. Fan, and W. Chen, "A hierarchical neural network for sleep stage classification based on comprehensive feature learning and multi-flow sequence learning," *IEEE journal of biomedical and health informatics*, vol. 24, no. 5, pp. 1351–1366, 2019.

[23] A. Supratak and Y. Guo, "Tinysleepnet: An efficient deep learning model for sleep stage scoring based on raw single-channel eeg," in *2020 42nd Annual International Conference of the IEEE Engineering in Medicine & Biology Society (EMBC)*. IEEE, 2020, pp. 641–644.

[24] H. Korkalainen, J. Aakko, S. Nikkonen, S. Kainulainen, A. Leino, B. Duce, I. O. Afara, S. Myllymaa, J. Toyras, and T. Leppanen, "Accurate Deep Learning-Based Sleep Staging in a Clinical Population with Suspected Obstructive Sleep Apnea," *IEEE Journal of Biomedical and Health Informatics*, vol. 24, no. 7, pp. 2073–2081, 2020.

[25] M. Perslev, M. Jensen, S. Darkner, P. J. Jennum, and C. Igel, "U-time: A fully convolutional network for time series segmentation applied to sleep staging," *Advances in Neural Information Processing Systems*, vol. 32, pp. 4415–4426, 2019.

[26] Z. Jia, Y. Lin, J. Wang, X. Wang, P. Xie, and Y. Zhang, "Salientsleep-net: Multimodal salient wave detection network for sleep staging," *arXiv preprint arXiv:2105.13864*, 2021.

[27] H. Phan, K. Mikkelsen, O. Y. Chén, P. Koch, A. Mertins, and M. D. Vos, "Sleeptransformer: Automatic sleep staging with interpretability and uncertainty quantification," 2021.

[28] W. Qu, Z. Wang, H. Hong, Z. Chi, D. D. Feng, R. Grunstein, and C. Gordon, "A residual based attention model for eeg based sleep staging," *IEEE journal of biomedical and health informatics*, vol. 24, no. 10, pp. 2833–2843, 2020.

[29] C. Sun, J. Fan, C. Chen, W. Li, and W. Chen, "A two-stage neural network for sleep stage classification based on feature learning, sequence learning, and data augmentation," *IEEE Access*, vol. 7, pp. 109 386–109 397, 2019.

[30] J. Huang, L. Ren, X. Zhou, and K. Yan, "An improved neural network based on senet for sleep stage classification," *IEEE Journal of Biomedical and Health Informatics*, 2022.

[31] L. Fiorillo, P. Favaro, and F. D. Faraci, "DeepSleepNet-Lite: A Simplified Automatic Sleep Stage Scoring Model with Uncertainty Estimates," *IEEE Transactions on Neural Systems and Rehabilitation Engineering*, vol. 29, pp. 2076–2085, 2021.

[32] H. Wang, C. Lu, Q. Zhang, Z. Hu, X. Yuan, P. Zhang, and W. Liu, "A novel sleep staging network based on multi-scale dual attention," *Biomedical Signal Processing and Control*, vol. 74, p. 103486, 2022.

[33] M. N. Mohsenvand, M. R. Izadi, and P. Maes, "Contrastive representation learning for electroencephalogram classification," in *Machine Learning for Health*. PMLR, 2020, pp. 238–253.

[34] X. Jiang, J. Zhao, B. Du, and Z. Yuan, "Self-supervised Contrastive Learning for EEG-based Sleep Staging," *Proceedings of the International Joint Conference on Neural Networks*, vol. 2021-July, 2021.

[35] J. Ye, Q. Xiao, J. Wang, H. Zhang, J. Deng, and Y. Lin, "Cosleep: A multi-view representation learning framework for self-supervised learning of sleep stage classification," *IEEE Signal Processing Letters*, vol. 29, pp. 189–193, 2021.

[36] M. Caron, I. Misra, J. Mairal, P. Goyal, P. Bojanowski, and A. Joulin, "Unsupervised learning of visual features by contrasting cluster assignments," *Advances in Neural Information Processing Systems*, vol. 33, pp. 9912–9924, 2020.

[37] K. He, H. Fan, Y. Wu, S. Xie, and R. Girshick, "Momentum contrast for unsupervised visual representation learning," in *Proceedings of the IEEE/CVF conference on computer vision and pattern recognition*, 2020, pp. 9729–9738.

[38] T. Chen, S. Kornblith, M. Norouzi, and G. Hinton, "A simple framework for contrastive learning of visual representations," in *International conference on machine learning*. PMLR, 2020, pp. 1597–1607.

[39] R. Hadsell, S. Chopra, and Y. LeCun, "Dimensionality reduction by learning an invariant mapping," in *2006 IEEE Computer Society Conference on Computer Vision and Pattern Recognition (CVPR'06)*, vol. 2. IEEE, 2006, pp. 1735–1742.

[40] F. Schroff, D. Kalenichenko, and J. Philbin, "Facenet: A unified embedding for face recognition and clustering," in *Proceedings of the IEEE conference on computer vision and pattern recognition*, 2015, pp. 815–823.

[41] K. Sohn, "Improved deep metric learning with multi-class n-pair loss objective," *Advances in neural information processing systems*, vol. 29, 2016.

[42] T.-Y. Lin, P. Dollár, R. Girshick, K. He, B. Hariharan, and S. Belongie, "Feature pyramid networks for object detection," in *Proceedings of the IEEE conference on computer vision and pattern recognition*, 2017, pp. 2117–2125.

[43] P. Khosla, P. Teterwak, C. Wang, A. Sarna, Y. Tian, P. Isola, A. Maschinot, C. Liu, and D. Krishnan, "Supervised contrastive learning," *Advances in Neural Information Processing Systems*, vol. 33, 2020.

[44] A. L. Goldberger, L. A. Amaral, L. Glass, J. M. Hausdorff, P. C. Ivanov, R. G. Mark, J. E. Mietus, G. B. Moody, C.-K. Peng, and H. E. Stanley, "Physiobank, physiotoolkit, and physionet: components of a new research resource for complex physiologic signals," *circulation*, vol. 101, no. 23, pp. e215–e220, 2000.

[45] B. Kemp, A. H. Zwinderman, B. Tuk, H. A. Kamphuisen, and J. J. Oberye, "Analysis of a sleep-dependent neuronal feedback loop: the slow-wave microcontinuity of the eeg," *IEEE Transactions on Biomedical Engineering*, vol. 47, no. 9, pp. 1185–1194, 2000.

[46] K. Hermann, T. Chen, and S. Kornblith, "The origins and prevalence of texture bias in convolutional neural networks," in *Advances in Neural Information Processing Systems*, H. Larochelle,

M. Ranzato, R. Hadsell, M. Balcan, and H. Lin, Eds., vol. 33.   Curran Associates, Inc., 2020, pp. 19 000–19 015. [Online]. Available: https://proceedings.neurips.cc/paper/2020/file/db5f9f42a7157abe65bb145000b5871a-Paper.pdf

[47] K. He, X. Zhang, S. Ren, and J. Sun, "Delving deep into rectifiers: Surpassing human-level performance on imagenet classification," in *Proceedings of the IEEE international conference on computer vision*, 2015, pp. 1026–1034.

[48] J. Hu, L. Shen, and G. Sun, "Squeeze-and-excitation networks," in *Proceedings of the IEEE conference on computer vision and pattern recognition*, 2018, pp. 7132–7141.

[49] A. Vaswani, N. Shazeer, N. Parmar, J. Uszkoreit, L. Jones, A. N. Gomez, Ł. Kaiser, and I. Polosukhin, "Attention is all you need," in *Advances in neural information processing systems*, 2017, pp. 5998–6008.

[50] S. Hochreiter and J. Schmidhuber, "Long short-term memory," *Neural computation*, vol. 9, no. 8, pp. 1735–1780, 1997.

[51] J. Chung, C. Gulcehre, K. Cho, and Y. Bengio, "Empirical evaluation of gated recurrent neural networks on sequence modeling," *arXiv preprint arXiv:1412.3555*, 2014.

[52] G. Shi, Z. Chen, and R. Zhang, "A transformer-based spatial-temporal sleep staging model through raw eeg," in *2021 International Conference on High Performance Big Data and Intelligent Systems (HPBD&IS)*.   IEEE, 2021, pp. 110–115.

[53] M.-T. Luong, H. Pham, and C. D. Manning, "Effective approaches to attention-based neural machine translation," *arXiv preprint arXiv:1508.04025*, 2015.

[54] D. Bahdanau, K. Cho, and Y. Bengio, "Neural machine translation by jointly learning to align and translate," *arXiv preprint arXiv:1409.0473*, 2014.

[55] T. Hastie, R. Tibshirani, and J. Friedman, "The elements of statistical learning. springer series in statistics," *New York, NY, USA*, 2001.

[56] C. O'reilly, N. Gosselin, J. Carrier, and T. Nielsen, "Montreal archive of sleep studies: an open-access resource for instrument benchmarking and exploratory research," *Journal of sleep research*, vol. 23, no. 6, pp. 628–635, 2014.

[57] M. M. Ghassemi, B. E. Moody, L.-W. H. Lehman, C. Song, Q. Li, H. Sun, R. G. Mark, M. B. Westover, and G. D. Clifford, "You snooze, you win: the physionet/computing in cardiology challenge 2018," in *2018 Computing in Cardiology Conference (CinC)*, vol. 45.   IEEE, 2018, pp. 1–4.

[58] G.-Q. Zhang, L. Cui, R. Mueller, S. Tao, M. Kim, M. Rueschman, S. Mariani, D. Mobley, and S. Redline, "The national sleep research resource: towards a sleep data commons," *Journal of the American Medical Informatics Association*, vol. 25, no. 10, pp. 1351–1358, 2018.

[59] S. F. Quan, B. V. Howard, C. Iber, J. P. Kiley, F. J. Nieto, G. T. O'Connor, D. M. Rapoport, S. Redline, J. Robbins, J. M. Samet *et al.*, "The sleep heart health study: design, rationale, and methods," *Sleep*, vol. 20, no. 12, pp. 1077–1085, 1997.

[60] J. Long, E. Shelhamer, and T. Darrell, "Fully convolutional networks for semantic segmentation," in *Proceedings of the IEEE conference on computer vision and pattern recognition*, 2015, pp. 3431–3440.

[61] O. Ronneberger, P. Fischer, and T. Brox, "U-net: Convolutional networks for biomedical image segmentation," in *International Conference on Medical image computing and computer-assisted intervention*.   Springer, 2015, pp. 234–241.

[62] D. P. Kingma and J. Ba, "Adam: A method for stochastic optimization," *arXiv preprint arXiv:1412.6980*, 2014.

[63] A. Paszke, S. Gross, F. Massa, A. Lerer, J. Bradbury, G. Chanan, T. Killeen, Z. Lin, N. Gimelshein, L. Antiga *et al.*, "Pytorch: An imperative style, high-performance deep learning library," *Advances in neural information processing systems*, vol. 32, 2019.

[64] M. Sokolova and G. Lapalme, "A systematic analysis of performance measures for classification tasks," *Information processing & management*, vol. 45, no. 4, pp. 427–437, 2009.

[65] R. Geirhos, P. Rubisch, C. Michaelis, M. Bethge, F. A. Wichmann, and W. Brendel, "Imagenet-trained CNNs are biased towards texture; increasing shape bias improves accuracy and robustness." in *International Conference on Learning Representations*, 2019. [Online]. Available: https://openreview.net/forum?id=Bygh9j09KX

[66] N. Carion, F. Massa, G. Synnaeve, N. Usunier, A. Kirillov, and S. Zagoruyko, "End-to-end object detection with transformers," in *European conference on computer vision*.   Springer, 2020, pp. 213–229.

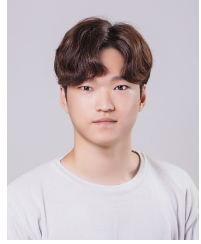

**Seongju Lee** received the B.S. degree in mechanical engineering from the Gwangju Institute of Science and Technology (GIST), South Korea, in 2018, and the M.S. degree in intelligent robotics from the School of Integrated Technology, GIST, where he is currently pursuing the Ph.D. degree with the School of Integrated Technology, under intelligent robotics program. His research interest is deep learning for health care, especially automatic sleep stage classification.

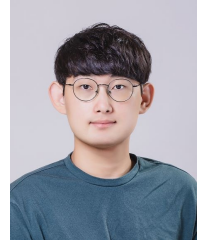

**Yeonguk Yu** received the B.S. degree in computer engineering from Hanbat National University, Daejeon, South Korea, in 2020. He received the M.S. degree from the Gwangju Institute of Science and Technology (GIST). He is currently pursuing the Ph.D. degree in intelligent robotics with the School of Integrated Technology, GIST. His current research interest is out-of-distribution data detection for deep learning applications.

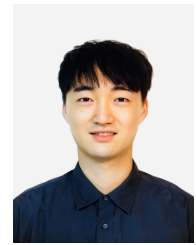

**Seunghyeok Back** (Graduate Student Member, IEEE) received the B.S. degree in mechanical engineering from the Gwangju Institute of Science and Technology (GIST), Gwangju, South Korea, in 2018, where he is currently pursuing the Ph.D. degree with the School of Integrated Technology, under the intelligent robotics program. His current research interests are deep learning for robotic manipulation and health care application.

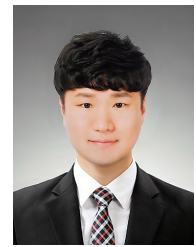

**Hogeon Seo** received the Ph.D. degree in convergence mechanical engineering from Hanyang University in 2018. He is a Senior Researcher in the Artificial Intelligence Application & Strategy Team of Korea Atomic Energy Research Institute. His research interests are artificial intelligence for prognosis, nondestructive evaluation, and sensor fusion.

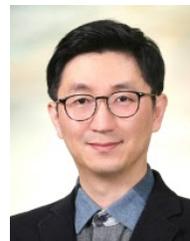

**Kyoobin Lee** (Member, IEEE) received the B.S., M.S., and Ph.D. degrees in mechanical engineering from the Korea Advanced Institute of Science and Technology (KAIST), South Korea, in 2008. From 2008 to 2010, he was a Postdoctoral Scholar with the Center for Neuroscience, KIST. From 2012 to 2017, he was a Principal Researcher with the Samsung Advanced Institute of Technology. From 2017 to 2021, he was an Assistant Professor with the School of Integrated Technology, Gwangju Institute of Science and Technology (GIST), where he has been an Associate Professor since 2022. His research interests include vision recognition using deep learning, robot control application using computer vision, and split learning for neural networks of cloud computing applications. He is the Editor of the Korea Robotics Society Review.